\documentclass[10pt,twocolumn,letterpaper]{article}

\usepackage{cvpr}
\usepackage{times}
\usepackage{epsfig}
\usepackage{graphicx}
\usepackage{amsmath}
\usepackage{amssymb}

\usepackage{epstopdf}
\usepackage{amsmath}
\usepackage{color}
\usepackage{cite}
\usepackage{lineno}
\usepackage{algorithmic} 
\usepackage{algorithm}
\usepackage{setspace}
\usepackage{subfigure}
\usepackage[normalem]{ulem}

\usepackage[pagebackref=true,breaklinks=true,letterpaper=true,colorlinks,bookmarks=false]{hyperref}

\cvprfinalcopy 

\def\cvprPaperID{****} 

\ifcvprfinal\fi
\begin{document}

\title{Bidirectional Warping of Active Appearance Model}

\author{Ali Mollahosseini and Mohammad H. Mahoor\\
Department of Electrical and Computer Engineering\\
University of Denver, Denver, CO 80210\\
\tt\small{ali.mollahosseini@du.edu and mmahoor@du.edu}
}

\maketitle

\begin{abstract}
Active Appearance Model (AAM) is a commonly used method for facial image analysis with applications in face identification and facial expression recognition. This paper proposes a new approach based on image alignment for AAM fitting called bidirectional warping. Previous approaches warp either the input image or the appearance template. We propose to warp both the input image, using incremental update by an affine transformation, and the appearance template, using an inverse compositional approach. Our experimental results on Multi-PIE face database show that the bidirectional approach outperforms state-of-the-art inverse compositional fitting approaches in extracting landmark points of faces with shape and pose variations.

\end{abstract}

\section{Introduction}

Facial landmark point extraction is a key step in facial image representation and analysis. The Active Appearance Model (AAM) proposed by \textit{Cootes et al.}~\cite{Cootes1998AAM} is a powerful object description method that is commonly used for facial landmark points extraction~\cite{Cootes1998AAM, matthews2004active}, facial action unit extraction~\cite{mahoor2009framework}, medical image segmentation and analysis~\cite{cootes2001statistical}. The idea behind AAM is to represent a visual object (e.g. facial image) using a linear model of shape and texture (appearance) eigenvectors obtained from a set of manually labeled training images. Then, the model is used to represent an instance of the object in a novel image. This process is often called \textbf{AAM fitting}. 

AAM fitting is a non-linear optimization problem. Different optimization approaches have been proposed to find the best model parameters that result in minimum error between the synthesized appearance models obtained from the AAM and the input image. In general, due to variation of camera view angle, resolution and focal distance, facial images have different scaling, rotation, and translations. In order to remove global shape variations, all shapes are normalized and the modeling is only concerned with local shape deformation. Therefore, it is necessary to combine a global shape transformation with the normalized AAM. The global shape transformation is often a 2D similarity transformation. Finding optimal parameters of the global transformation improves the accuracy of fitting in representing novel facial images with different shape and pose variations.

Traditionally, the stochastic gradient descent algorithm or iteratively incremental additive techniques are used to update the AAM parameters to fit onto novel images~\cite{Cootes1998AAM}. The fitting problem can also be viewed as finding a model instance similar to the given facial image and therefore it can be considered as an image alignment problem. Baker and Matthews~\cite{baker2004lucas} have categorized these approaches into four classes: Forwards Additive, Forwards Compositional, Inverse Additive, and Inverse Compositional. They proposed the Projecting Out (PO) technique which is admittedly one of the fastest algorithms for AAM fitting~\cite{matthews2004active}. They also proposed the Simultaneously Inverse Compositional (SIC) method that can handle images of subjects not included in the training better at the price of losing speed~\cite{Gross2005}. 

In the literature, there are some works~\cite{Keller2004fast, Merget2010BidirectionalComposition} on image alignment for applications, such as motion estimation~\cite{Keller2004fast}, that take advantage of the gradients of both the template and target images. These approaches are called bidirectional image alignment. Bidirectional approaches work better than unidirectional image alignment approaches~\cite{Merget2010BidirectionalComposition}. In this paper, we reformulate AAM fitting using a bidirectional image alignment scheme. 

In our approach, we minimize the error between a warped image and the appearance template by iteratively solving a non-linear least square problem. The warping is a piecewise affine of a normalized AAM that is followed by a global transformation. In each iteration, shape parameters are optimized based on the trained appearance template using the Inverse Compositional Algorithm (ICA)~\cite{baker2004lucas}, and global transformation is found based on the gradient of the input image using incremental update. We call this approach bidirectional warping. Moreover, we utilize affine transformation instead of 2D similarity to increase the generality of the global shape transformation, and apply a fitting constraint to prevent the algorithm from resulting in non-face shapes. We show that the proposed bidirectional approach can be applied to PO and SIC fitting methods. We study the performance of the proposed bidirectional PO and SIC methods in extracting facial landmark points, and examine and compare the effect of proposed affine transformation, and the fitting constraint on both bidirectional and the original PO and SIC fitting methods.

The rest of this paper is organized as follows. Section~\ref{BACKGROUND} briefly introduces AAM algorithms and particularly reviews image alignment-based AAM fitting. Section~\ref{Bidirectional_Warping_AAM_Fitting} describes the bidirectional warping method. Experimental results are given in Section~\ref{ExperimentalResults}, and Section~\ref{Discussion_Conclusions} concludes the paper.

\section{BACKGROUND}
\label{BACKGROUND}

AAM consists of a shape component and an appearance component obtained from a set of annotated landmark points in training images. Let's assume we are given a training facial image set with annotated shapes defined as: $\textbf{s} = (x_1, y_1, x_2, y_2,..., x_v, y_v)^T$. The training images are first normalized and aligned using iterative Procrustes analysis~\cite{cootes2001statistical}. This step removes variations due to a chosen global shape normalization transformation so that the resulting model can efficiently consider local and non-rigid shape deformation. We then can combine the resulting AAM with a global transformation. Afterwards, Principal Component Analysis (PCA) is applied to the set of normalized training shapes and a shape model is defined as:
\begin{equation}
\label{eq:PCAShapeModeling}
\textbf{s} = \textbf{s}_0 + \sum_{i=1}^{n}{p_i\textbf{s}_i},
\end{equation}
where the base shape $\textbf{s}_0$ is the mean shape and the vectors $\textbf{s}_i$ are $n$ eigenvectors corresponding to the $n$ largest eigenvalues. Then, all the training images are normalized by warping them into the base shape $\textbf{s}_0$, using piecewise affine warp, and the appearance model is defined as:
\begin{equation}
\label{eq:PCA-AppearanceModeling}
\mbox{A}(\textbf{x}) = \mbox{A}_0(\textbf{x}) + \sum_{i=1}^{m}{\lambda_i \mbox{A}_i(\textbf{x})} \qquad \forall\mbox{x} \in \textbf{s}_0,
\end{equation}
where $ \mbox A_0$ is the mean appearance and the vectors $ \mbox A_i$ are the $m$ eigenvectors corresponding to the $m$ largest eigenvalues.

The goal of fitting is to find a model instance that can efficiently describe the object (e.g. face) in a given image. Thus, it can be considered as an image alignment problem. In other words, we want to find the model instance $M(\textbf{W}(\textbf{x}; \textbf{p})) = A(\textbf{x})$ as similar as the image $I(\textbf{x})$. 

In general, facial images have different scaling, rotation, and translations. Therefore, it is necessary to combine a global shape transformation with the normalized AAM. If we consider the global shape transformation as $\textbf{N}\left(\textbf{x};\textbf{q}\right)$, we want to minimize the error between the template and $I\left(\textbf{N}\left(\textbf{W}\left(\textbf{x};\textbf{p}\right);\textbf{q}\right)\right)$. Considering global shape transformation, the objective of the fitting process is to find $\textbf{p}$ and $\textbf{q}$ in order to minimize the error image as:
\begin{equation}
\label{eq:ErrorImageAAMFitting}
E(\mbox{x})=\sum_{\mbox{x}\in \textbf{s}_0}{ \left[  \mbox A_0 (\textbf{x}) - I\left(\textbf{N}\left(\textbf{W}\left(\textbf{x};\textbf{p}\right);\textbf{q}\right)\right)\right]^2},
\end{equation}
which is a non-linear least square problem. We can have different definitions for the global transformation $\textbf{N}\left(\textbf{x};\textbf{q}\right)$. In~\cite{matthews2004active}, a set of 2D similarity transformations as a subset of piecewise affine warps is defined. Assuming the base mesh $\textbf{s}_0 = \left(x_1^0, y_1^0,..., x_v^0, y_v^0\right)^\textrm{T}$, we choose $\textbf{s}_1^{\ast} = \textbf{s}_0$, $\textbf{s}_2^{\ast} = \left(-y_1^0,x_1^0,..., -y_v^0,x_v^0\right)^\textrm{T}$, $\textbf{s}_3^{\ast}=\left(1,0,\dots,1,0\right)^\textrm{T}$ and $\textbf{s}_4^{\ast}=\left(0,1,\dots,0,1\right)^\textrm{T}$, then global transformation is $\textbf{N}\left(\textbf{x};\textbf{q}\right)= \textbf{s}_0 + \sum_{i=1}^{4}{q_i \textbf{s}_i^{\ast}}$. This representation of $\textbf{N}\left(\textbf{x};\textbf{q}\right)$ is similar to $\textbf{W}\left(\textbf{x};\textbf{p}\right)$ and therefore similar analysis on the shape parameters \textbf{p} can be applied to \textbf{q}. If we assume that the two sets of shape vectors $\textbf{s}_i$ and $\textbf{s}_i^{\ast}$ are orthogonal to each other, we can add the four 2D similarity vectors $\textbf{s}_i^{\ast}$ to the beginning of AAM shape vectors $\textbf{s}_i$~\cite{matthews2004active} and model any given shape as: $\textbf{s} = \textbf{s}_0 + \sum_{i=1}^{n+4}{p_i\textbf{s}_i}$. In practice, $\textbf{s}_i$ and $\textbf{s}_i^{\ast}$ are not quite orthogonal to each other. This can either be ignored when the size of $\textbf{s}_i$ is small or the complete set of $\textbf{s}_i$ and $\textbf{s}_i^{\ast}$ can be orthonormalized preferably. 

In~\cite{baker2004lucas}, \textit{Baker et al.} relate AAM to the Lucas-Kanade algorithm. They proposed the Inverse Compositional Algorithm (ICA), in which they find shape variation on the template and compose the inverse of that with the current shape. Therefore, many computationally expensive tasks are precomputed. 

In~\cite{matthews2004active}, appearance variation is considered in the fitting by finding shape parameters in a linear subspace where the appearance variation is ignored and then ``projected out'' to the full space with respect to the appearance eigenvectors. The method is more generic compared with the ICA, but the fitting is not accurate when applied to subjects that are not similar to subjects in the training set. The ``projecting out'' approach is called PO in the rest of this paper.

In~\cite{Gross2005}, Simultaneously Inverse Compositional (SIC) method is introduced, which is more generic. In this method the fitting procedure minimizes the error between $\left[  \mbox A_0 (\mbox{x}) + \sum_{i=1}^{m}{\left(\lambda_i + \Delta \lambda_i\right) A_i}\right]$ and $I\left(\textbf{N}\left(\textbf{W}\left(\textbf{x;p}\right);\textbf{q}\right)\right)$, where $ \mbox A_i$ are $m$ appearance eigenvectors correspond to the $m$ largest appearance eigenvalues, and $\left(\lambda_i + \Delta \lambda_i\right)$ are parameters of appearance that are found simultaneously with respect to the $\Delta \textbf{p}$. As the appearance parameters are optimized in each iteration, both steepest descent and the Hessian matrix $\left(H\right)$ should be calculated in each iteration, and therefore the method is slower. In~\cite{Gross2005} the PO is compared with the SIC, and the SIC is reported more accurate in modeling unseen subjects.

\section{Bidirectional Warping for AAM Fitting}
\label{Bidirectional_Warping_AAM_Fitting}

In this paper, we optimize the global transformation's parameters ($\textbf{q}$) based on $I$, using an incremental update and the shape's parameters ($\textbf{p}$) based on $ \mbox A_0$, using inverse compositional approach. If we assume $\textbf{p}$ and $\textbf{q}$ are known, reversing the role of $\textbf{W}$ in $I\left(\textbf{N}\left(\textbf{W}\left(\textbf{x};\textbf{p}\right);\textbf{q}\right)\right)$ and computing the incremental global warp $\textbf{N}$ with respect to $\textbf{W}$ in $I\left(\textbf{N}\left(\textbf{W}\left(\textbf{x};\textbf{p}\right);\textbf{q}\right)\right)$, we can solve the Equation \eqref{eq:ErrorImageAAMFitting} iteratively as:
 \begin{equation}
 \label{eq:IncrementalDeltaErrorForm}
  \footnotesize{\sum_{\mbox{x}\in \textbf{s}_0}{ \left[  \mbox A_0 \left(\textbf{N}\left(\textbf{W}\left(\textbf{x};0+\Delta\textbf{p}\right);0\right)\right) - I\left(\textbf{N}\left(\textbf{W}\left(\textbf{x};\textbf{p}\right);\textbf{q}+\Delta\textbf{q}\right)\right)\right]^2}}.
 \end{equation}

Then to update the warping parameters, we use $\textbf{W}(\textbf{x}; \textbf{p}) \leftarrow \textbf{W}(\textbf{x}; \textbf{p}) \circ \textbf{W}(\textbf{x}; \Delta \textbf{p})^{-1}$ and $\textbf{q}=\textbf{q}+\Delta \textbf{q}$. Assuming $\textbf{W}\left(\textbf{x};0\right)$ and $\textbf{N}\left(\textbf{x};0\right)$ are identity warps, first order Taylor series expansion of the Equation \eqref{eq:IncrementalDeltaErrorForm} on $\Delta\textbf{p}$ and $\Delta\textbf{q}$ gives:
    \begin{equation}
    \label{eq:TaylorExpansionPQ}
    \footnotesize{\sum_{\mbox{x}\in \textbf{s}_0}{ \left[  \mbox A_0 + \nabla  \mbox A_0 \frac{\partial \mbox{\textbf{W}}}{\partial \textbf{p}} \Delta \textbf{p} - I\left(\textbf{N}\left(\textbf{W}\left(\textbf{x;p}\right);\textbf{q}\right) \right)
    - \nabla I \frac{\partial \mbox{\textbf{N}}}{\partial \textbf{q}} \Delta \textbf{q} \right]^2}},
    \end{equation}
 \begin{normalsize}
 where $\nabla$ is the image gradient, $\frac{\partial \mbox{\textbf{W}}}{\partial \textbf{p}}$ and $\frac{\partial \mbox{\textbf{N}}}{\partial \textbf{q}}$ are the Jacobian of the warp evaluated at $\textbf{p}=0$ and current $\textbf{q}$ respectively. By taking the derivative of the Equation \eqref{eq:TaylorExpansionPQ}, neglecting second order $\Delta\textbf{p}\Delta\textbf{q}$ terms and optimizing for $\Delta\textbf{p}$ and $\Delta\textbf{q}$, we obtain:
 \end{normalsize}
\begin{small}
\begin{subequations}
\begin{flalign}
                  \Delta \textbf{p}= \mbox{H}_1^{-1}\sum_{\mbox{x}}{ \left[ \nabla  \mbox A_0 \frac{\partial \textbf{W}}{\partial \textbf{p}}\right]^T \left[I\left(\textbf{N}\left(\textbf{W}\left(\textbf{x};\textbf{p}\right);\textbf{q}\right) \right) -  \mbox A_0\right]}, \label{eq:deltaP}\\
                 \Delta \textbf{q}= \mbox{H}_2^{-1}\sum_{\mbox{x}}{ \left[ \mbox A_0 - I\left(\textbf{N}\left(\textbf{W}\left(\textbf{x};\textbf{p}\right);\textbf{q}\right) \right)\right] \left[ \nabla I \frac{\partial \textbf{N}}{\partial \textbf{q}}\right] }, \label{eq:deltaQ}
\end{flalign}
\end{subequations}
\end{small}
 \begin{normalsize}
where       
 \end{normalsize}
 \begin{small}
\begin{subequations}	
\begin{align}
                  \small{\mbox{H}_1=\left[ \nabla  \mbox A_0 \frac{\partial \textbf{W}}{\partial \textbf{p}}\right]^T\left[ \nabla  \mbox A_0 \frac{\partial \textbf{W}}{\partial \textbf{p}}\right]},                   
              	\label{eq:HessianMatrix1}\\
                 \small{\mbox{H}_2=\left[ \nabla I \frac{\partial \textbf{N}}{\partial \textbf{q}}\right]^T\left[ \nabla I \frac{\partial \textbf{N}}{\partial \textbf{q}}\right]}. \label{eq:HessianMatrix2}
\end{align}
\end{subequations}
\end{small}
 \begin{normalsize}
As $\small{\frac{\partial \mbox{\textbf{N}}}{\partial \textbf{q}}}$ is evaluated at $\textbf{p}=0$, $\mbox{H}_1$ can be precomputed and saved in the memory, while $\mbox{H}_2$ depends on the current shape and the warped input image gradient, and therefore it should be computed in each iteration.  Algorithm~\ref{alg:BidirectionalWarping} shows the steps of the bidirectional warping for inverse compositional algorithm. We call this approach Bi-ICA in the rest of this paper. 
 \end{normalsize}

\begin{algorithm}[h]
\caption{The Bidirectional Warping Algorithm}
\label{alg:BidirectionalWarping}
\renewcommand{\algorithmicrequire}{\textbf{Pre-compute:}}
\renewcommand{\algorithmicensure}{\textbf{Iterate:}}
\begin{algorithmic}[0]
\REQUIRE
\STATE (3) Evaluate the gradient $\nabla  \mbox A_0$ of the template $ \mbox A_0\left(\mbox{x}\right)$
\STATE (4) Evaluate the Jacobian $\frac{\partial \mbox{W}}{\partial \mbox{p}}$ at $\left(\mbox{x};0\right)$
\STATE (5) Compute the steepest descent images $\nabla  \mbox A_0 \frac{\partial \mbox{W}}{\partial \mbox{p}}$
\STATE (6) Compute the Hessian matrix $\mbox{H}_1$ using Equation \eqref{eq:HessianMatrix1}
\ENSURE
\STATE (1) Warp $I$ with $\mbox{W}\left(\mbox{x};\mbox{p}\right)$ and $\mbox{N}\left(\mbox{x};\mbox{q}\right)$ to compute\\
\hspace{0.5cm} $I\left(\mbox{N}\left(\mbox{W}\left(\mbox{x;p}\right);\mbox{q}\right) \right)$
\STATE (2) Compute $E=\left[I\left(\mbox{N}\left(\mbox{W}\left(\mbox{x;p}\right);\mbox{q}\right) \right) -  \mbox A_0\left(\mbox{x}\right)\right]$
\STATE (7) Evaluate the gradient $\nabla I\left(\mbox{N}\left(\mbox{W}\left(\mbox{x;p}\right);\mbox{q}\right) \right)$
\STATE (8) Evaluate the Jacobian $\frac{\partial \mbox{N}}{\partial \mbox{q}}$
\STATE (9) Compute the steepest descent images $\nabla I \frac{\partial \mbox{N}}{\partial \mbox{q}}$
\STATE (10) Compute the Hessian matrix $\mbox{H}_2$ using Equation \eqref{eq:HessianMatrix2}
\STATE (11) Compute $\Delta \mbox{p}$ and $\Delta \mbox{q}$ using Equation \eqref{eq:deltaP} and \eqref{eq:deltaQ}
\STATE (12) Update $\mbox{W}\left(\mbox{x};\mbox{p}\right) \leftarrow \mbox{W}\left(\mbox{x};\mbox{p}\right) \circ \mbox{W}\left(\mbox{x};- \Delta \mbox{p}\right)^{-1}$ \\
\hspace{0.6cm} and $\mbox{q}=\mbox{q}+ \Delta \mbox{q}$
\end{algorithmic}
\end{algorithm}

The ``projecting out'' technique can be applied to the bidirectional warping, i.e. instead of $\mbox{SD}=\left[ \nabla  \mbox A_0 \frac{\partial \mbox{W}}{\partial \mbox{p}}\right]$ in the Equation \eqref{eq:deltaP} and \eqref{eq:HessianMatrix1}, $\mbox{SD}$ is calculated as:
\begin{equation}
\label{eq:IncludingAppVarSteepestDescent}
\small{
\textrm{SD}(\mbox{x})=\nabla  \mbox A_0 \frac{\partial \mbox{W}}{\partial \mbox{p}} - \sum_{i=1}^{m}{\left[ \sum_{x \in \textbf{s}_0}{ \mbox A_i\left(\mbox{x}\right). \nabla  \mbox A_0 \frac{\partial \mbox{W}}{\partial \mbox{p}}} \right]  \mbox{A}_i(\mbox{x})}},
\end{equation}

Similar to the PO, the $H_1$ can be precomputed, but the dot product of the modified steepest descent images with the error image should be computed in each iteration. The bidirectional warping of the PO is called Bi-PO in the rest of this paper.

To have a more generic fitting, we can optimize the shape parameters on the full space of the appearance vectors. In this case, we need to optimize the appearance parameters as well as the shape parameters like the SIC method. The algorithm operates by iteratively minimizing:
\begin{align}
 f(x) &= \sum_{\mbox{x}\in \textbf{s}_0}\left[\mbox A_0 \left(\textbf{N}\left(\textbf{W}\left(\textbf{x};0+\Delta\textbf{p}\right);0\right)\right) \right.\nonumber\\
 &\qquad \left. {}   + \sum_{i=1}^{m}{\left(\lambda_i + \Delta \lambda_i\right) \mbox{A}_i\left(\textbf{N}\left(\textbf{W}\left(\textbf{x};0+\Delta\textbf{p}\right);0\right)\right)} \right.\nonumber\\
 &\qquad \left. {} - I\left(\textbf{N}\left(\textbf{W}\left(\textbf{x};\textbf{p}\right);\textbf{q}+\Delta\textbf{q}\right)\right)\right]^2,
\end{align}
simultaneously with respect to $\Delta\textbf{p}$, $\Delta\textbf{q}$ and $\Delta \lambda=(\Delta \lambda_1,...,\Delta \lambda_m)$. Then we update the warp $\textbf{W}(\textbf{x}; \textbf{p}) \leftarrow \textbf{W}(\textbf{x}; \textbf{p}) \circ \textbf{W}(\textbf{x}; \Delta \textbf{p})^{-1}$, $\textbf{q}=\textbf{q}+\Delta \textbf{q}$ and $\lambda=\lambda+\Delta \lambda$. 

We define the concatenation parameter of the shape and the appearance $\mbox{r}=[\mbox{p},\lambda]^T$, and the steepest-descent images as:
\begin{equation}
\label{eq:SICSD}
\small{\textrm{SD}_{sim}(\mbox{x})=\left( \nabla \mbox{A} \frac{\partial \mbox{W}}{\partial \mbox{p}_1},...,\nabla \mbox{A} \frac{\partial \mbox{W}}{\partial \mbox{p}_n}, \mbox{A}_1,...,\mbox{A}_m \right)}
\end{equation}
where $\nabla \mbox{A}=\nabla  \mbox A_0 + \sum_{i=1}^{m}{\lambda_i \nabla  \mbox{A}_i}$. We can then compute the parameter update $\Delta \mbox{r}$ as:
\begin{equation}
\label{eq:Delta_r}
\small{\Delta \mbox{r}= -\mbox{H}_{sim}^{-1}\sum_{\mbox{x}}{ \textrm{SD}_{sim}^T(\mbox{x}) E(\mbox{x})}}
\end{equation}
where $\mbox{H}_{sim}^{-1}= \sum_{\mbox{x}}{ \textrm{SD}_{sim}^T(\mbox{x}) \textrm{SD}_{sim}(\mbox{x})}$.

To find the parameter of the global transformation ($\textbf{q}$), we used incremental update as: $\textbf{q}=\textbf{q}+\Delta \textbf{q}$, where $\small{\Delta \textbf{q}= \mbox{H}_2^{-1}\sum_{\mbox{x}}{ -E(\mbox{x}) \left[ \nabla I \frac{\partial \textbf{N}}{\partial \textbf{q}}\right] }}$. This approach is called Bi-SIC in the rest of this paper. In this case both $\textrm{SD}_{sim}$ and $\mbox{H}_{sim}$ are calculated in each iteration. The extra computational load of Bi-SIC in comparison with SIC is to calculate the gradient of the warped image and $\mbox{H}_2$ in each iteration.

In addition to the introduced bidirectional approach, we also propose two modifications to AAM fitting as follows:

\textbf{1-Affine Transformation:} Image alignment techniques for AAM fitting usually consider a 2D set of similarity transform for the global transformation. Affine transformation can improve the performance of Active Shape Model for facial feature extraction~\cite{mahoor2006improved}. In this paper, we apply an affine transformation with six degrees of freedom for AAM fitting. Assuming the base mesh is: $\textbf{s}_0 = \left(x_1^0, y_1^0,..., x_v^0, y_v^0\right)^\textrm{T}$. We choose $\textbf{s}_1^{\ast} = \left(x_1^0,0,...,x_v^0,0\right)^\textrm{T}$, $\textbf{s}_2^{\ast} = \left(y_1^0,0,...,y_v^0,0\right)^\textrm{T}$, $\textbf{s}_3^{\ast} = \left(0,x_1^0,...,0,x_v^0\right)^\textrm{T}$, $\textbf{s}_4^{\ast} = \left(0,y_1^0,...,0,y_v^0\right)^\textrm{T}$,  $\textbf{s}_5^{\ast}=\left(1,0,\dots,1,0\right)^\textrm{T}$ and $\textbf{s}_6^{\ast}=\left(0,1,\dots,0,1\right)^\textrm{T}$. The global affine transformation is defined as: $\textbf{N}\left(\textbf{x};\textbf{q}\right)= \textbf{s}_0 + \sum_{i=1}^{6}{q_i \textbf{s}_i^{\ast}}$. This transformation has more degrees of freedom and therefore results in a better modeling of the shape variation.

\textbf{2- Fitting Constraint:} Introduced approaches for AAM fitting still suffer from lack of generality for unseen faces. In addition, the result can differ significantly from trained shapes. One idea is to apply some constraints on fitting iterations. Defining a well constraint is not easy because of the complexity of the face shape, huge variation of the appearance due to different subjects, illuminations and expressions, and the existence of non-face areas (e.g. glasses). In this paper, we apply a simple constraint of Active Shape Models (ASM)~\cite{cootes2001statistical}, i.e. those shape parameters ($\mbox{p}$) are updated that $p_i\leq 3\sqrt{b_i}$, where $b_i$ are the eigenvalues of the trained shapes. This constraint will force the algorithm to result in shapes similar to trained shapes with a limited degree of freedom and therefore prevent it from resulting in non-face shapes.

\section{Experimental Results}
\label{ExperimentalResults}
We implemented the PO~\cite{matthews2004active}, the SIC~\cite{Gross2005}, and our proposed Bi-PO and Bi-SIC methods using Matlab platform. We also used the affine transformation for the global transformation instead of 2D similarity and applied the introduced constraint to the PO, SIC, Bi-PO, and Bi-SIC methods and called them PO-AC, SIC-AC, Bi-PO-AC, and Bi-SIC-AC, respectively. 

\begin{figure}
        \centering
        \begin{subfigure}
                \centering
                \includegraphics[width=1in]{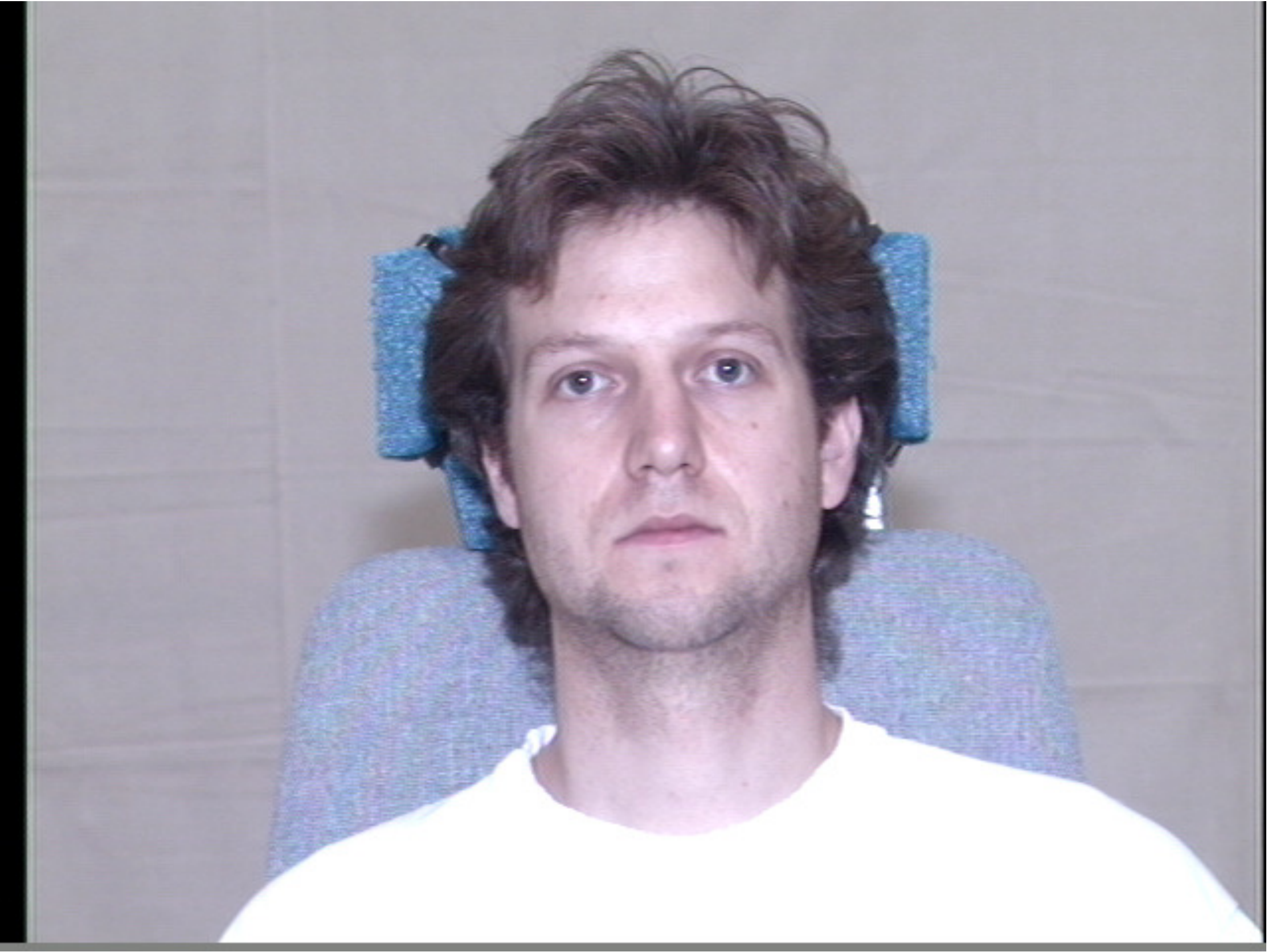}
        \end{subfigure}%
        ~ 
        \begin{subfigure}
                \centering
                \includegraphics[width=1in]{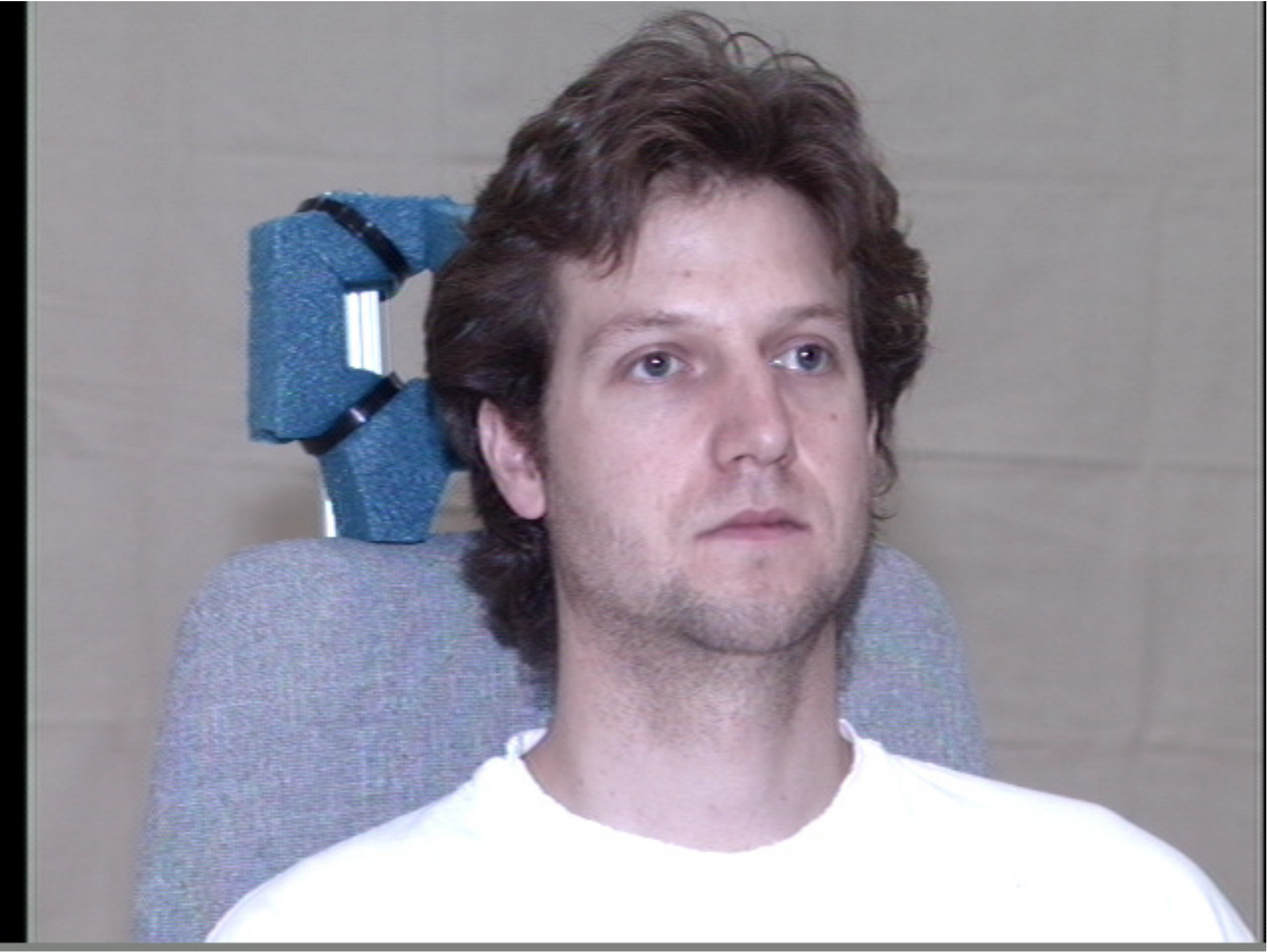}
        \end{subfigure}
        ~ 
        \begin{subfigure}
                \centering
                \includegraphics[width=1in]{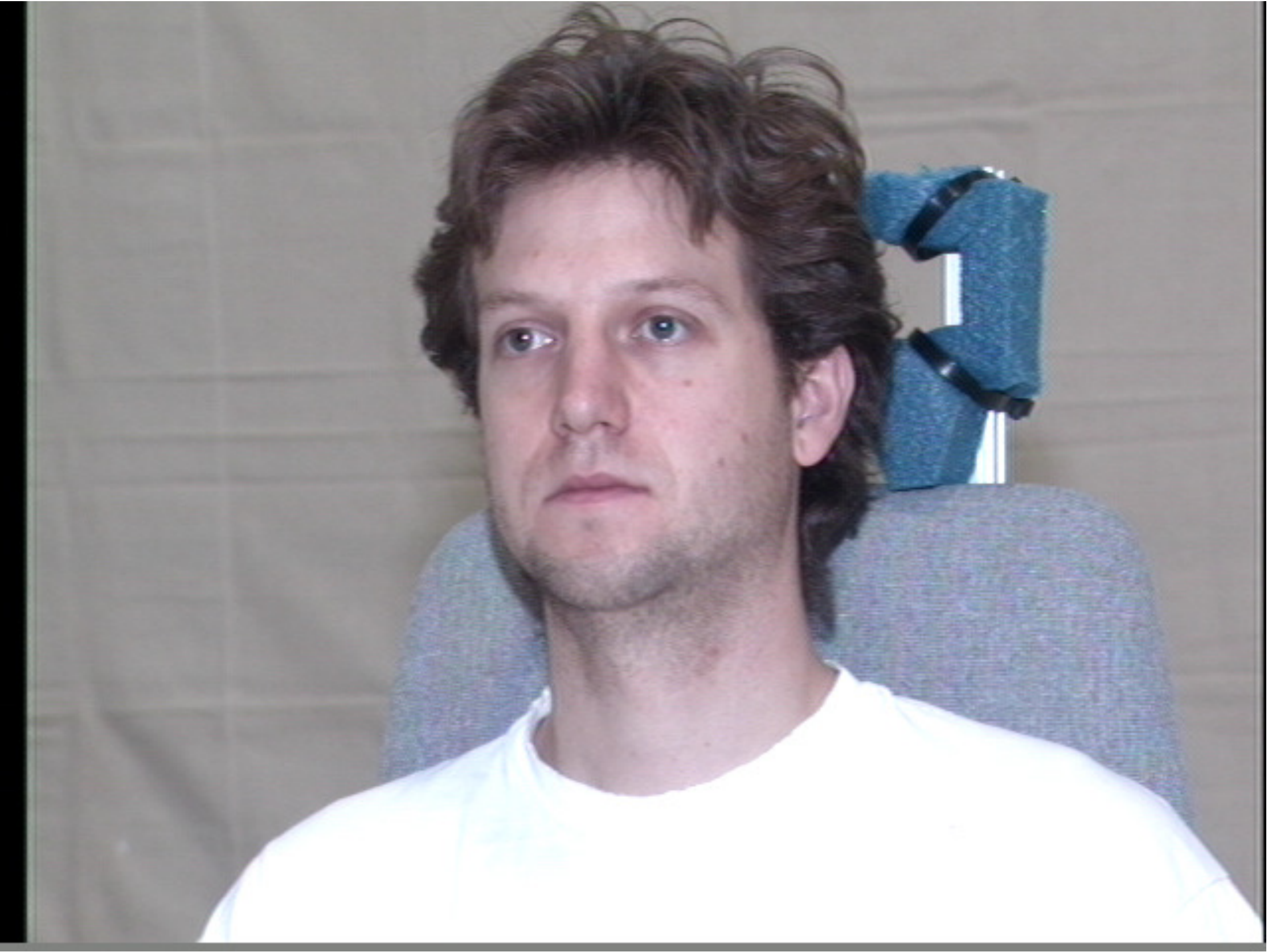}
        \end{subfigure}
        
        \begin{subfigure}
                \centering
                \includegraphics[width=1in]{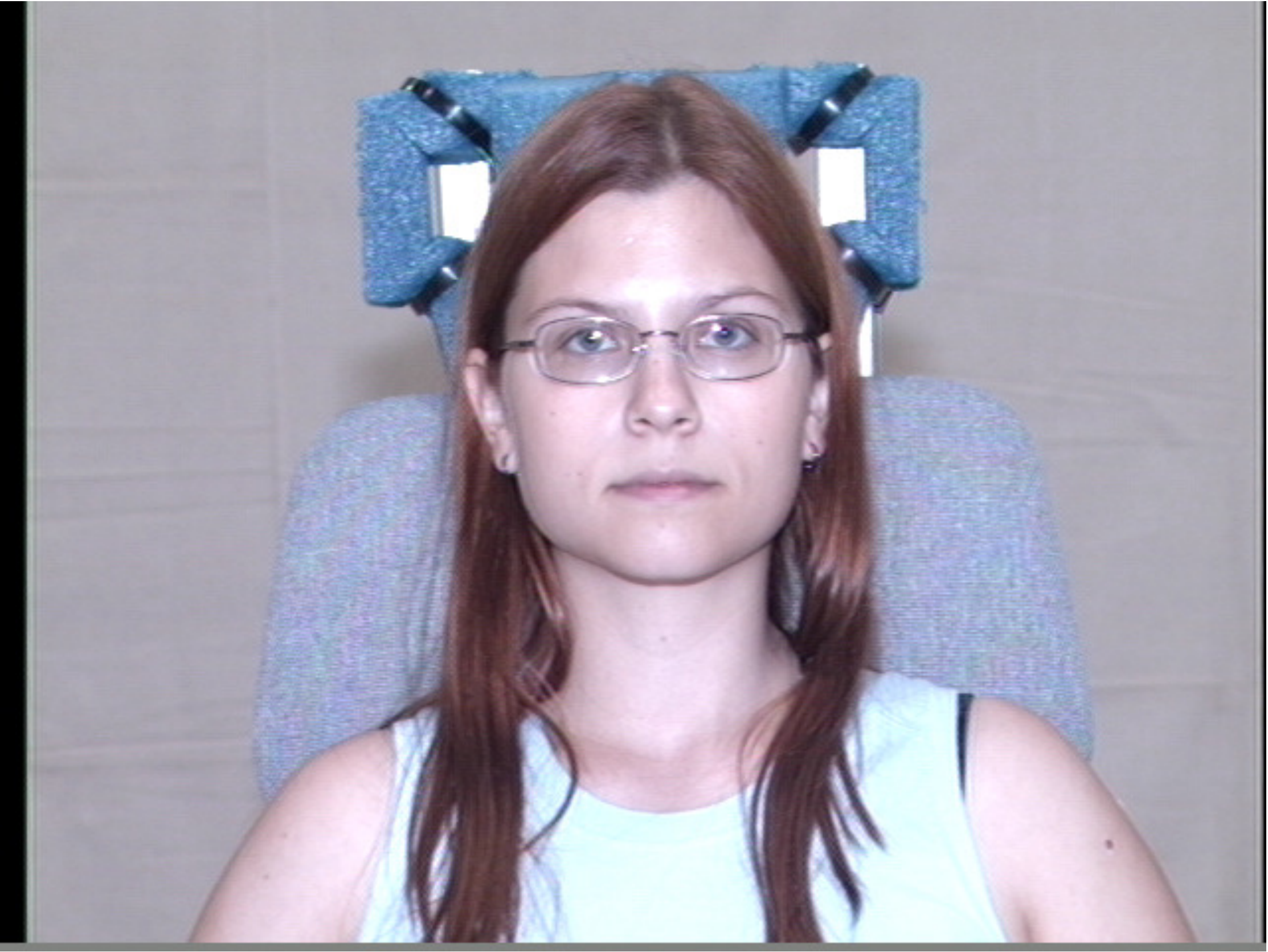}
        \end{subfigure}%
        ~ 
        \begin{subfigure}
                \centering
                \includegraphics[width=1in]{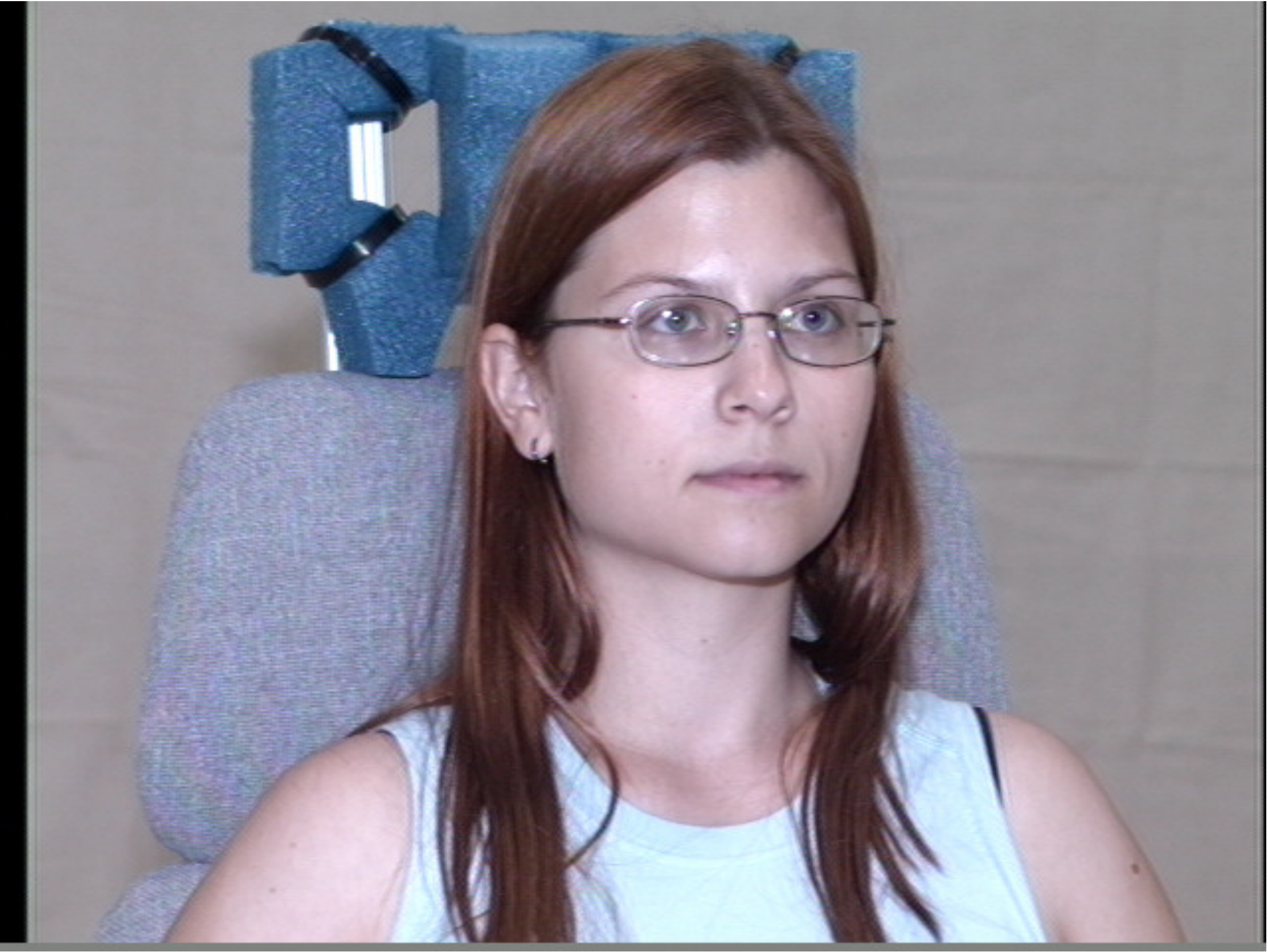}
        \end{subfigure}
        ~ 
        \begin{subfigure}
                \centering
                \includegraphics[width=1in]{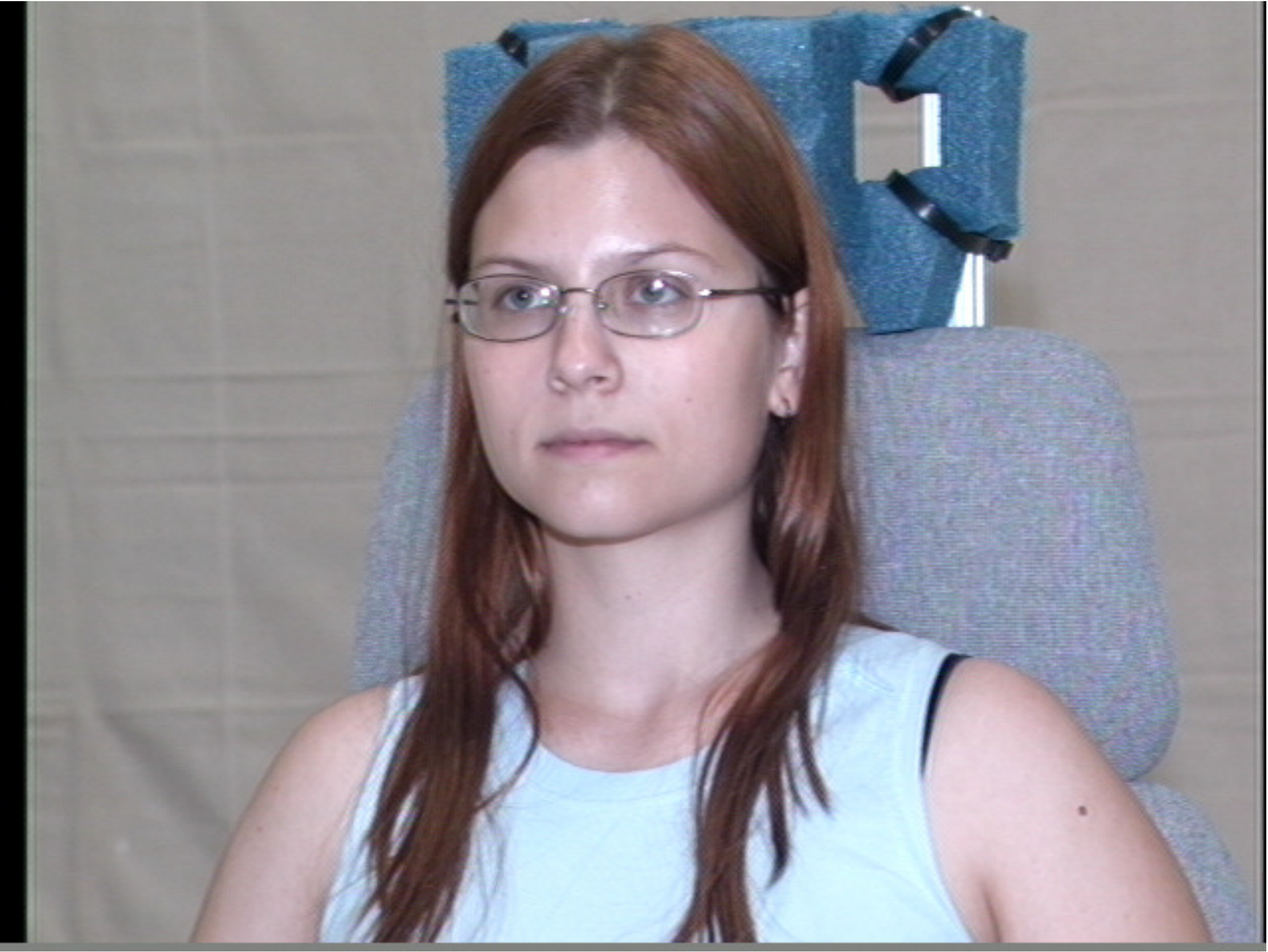}
        \end{subfigure}        
\caption{Some sample images of frontal, left and right poses from Multi-PIE dataset~\cite{MultiPIE}.}
\label{fig:SomeImages}
\vspace{-0.2cm}
\end{figure}

We applied the aforementioned methods on CMU Multi-PIE face dataset~\cite{MultiPIE}. The CMU Multi-PIE database contains more than 750,000 images of 337 people. Subjects were imaged under 15 view points and 19 illumination conditions. The image resolution is 640$\times$480, where the distance between the center of the eyes are approximately 80 pixels. Certain poses of a subset have 68 facial landmark points. We select a subset from the dataset containing 100 different subjects with the frontal head pose and with the same illumination. We also selected 50 images of left and right head poses that have 68 facial landmark points. Figure \ref{fig:SomeImages} shows images of sample subjects in frontal, left and right poses.

\begin{figure}[ht]
\vspace{-0.2cm}
\centering
 \subfigure[initial shape]{
   \includegraphics[width=1in]{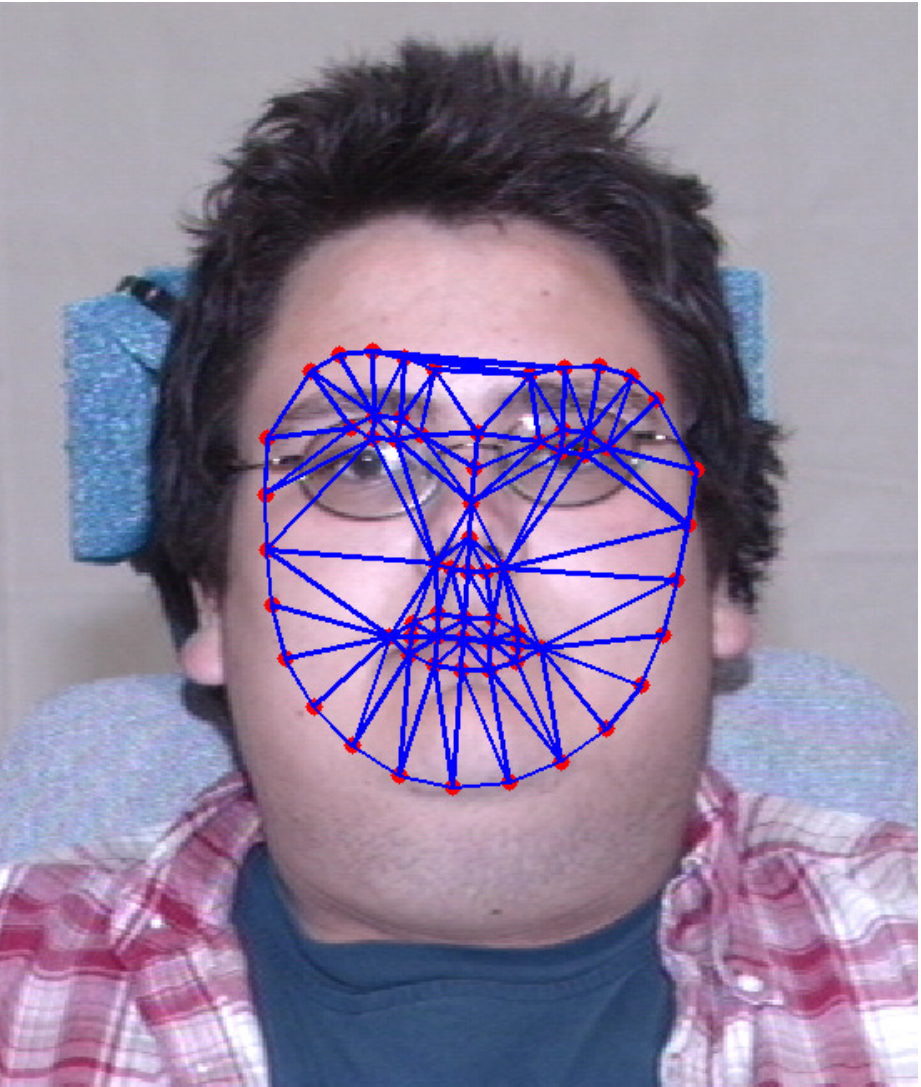}
   \label{fig:subfig_Initial}
 }
\subfigure[fitted shape]{
   \includegraphics[width=1in]{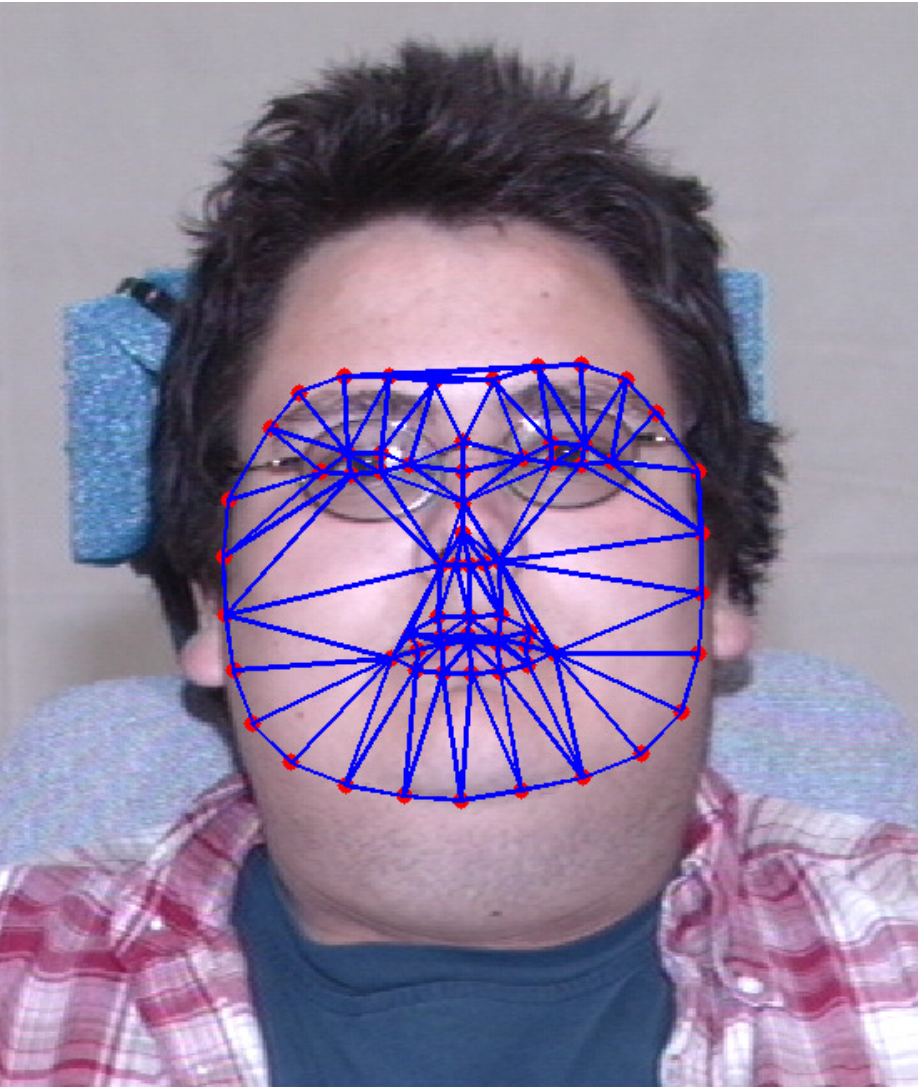}
   \label{fig:subfig_Fitted}
 }
\label{InitialFittedSample}
\caption{Initial and fitted shapes of a sample image.}
\vspace{-0.2cm}
\end{figure}

To initialize the shape model in AAM fitting, we selected two outer eye corners and the chin point (3 points) from the ground truth landmarks and perturbed them randomly by 5 pixels. Then we used the average shape obtained from training subjects as the initial shape and transformed it using similarity transformation obtained by those three perturbed points. Figure \ref{fig:subfig_Initial} shows the initial shape for a sample image.

We tested the performance of the PO, SIC, PO-AC, SIC-AC, Bi-PO, Bi-SIC, Bi-PO-AC, and Bi-SIC-AC methods when the number of images in the training sets varied using 10-fold cross validation. Particularly, we selected 10, 20, 30, 40, 50, 60, 70, 80 and 90 images randomly from the frontal subset and trained separate AAMs. For testing the generalization performance of the fitting methods, we fitted the trained models onto 10 images that are not included in the training sets and repeated this experiment 10 times for different test images. For comparing the fitting performance, we calculated the Root Mean Square Error (RMSE). The value of RMSE shows the distance between the fitted and the actual shape. Naturally, the smaller the RMSE, the better the fitting. 

\begin{figure}[!t]
\vspace{-0.2cm}
\centering
\includegraphics[width=2.4in]{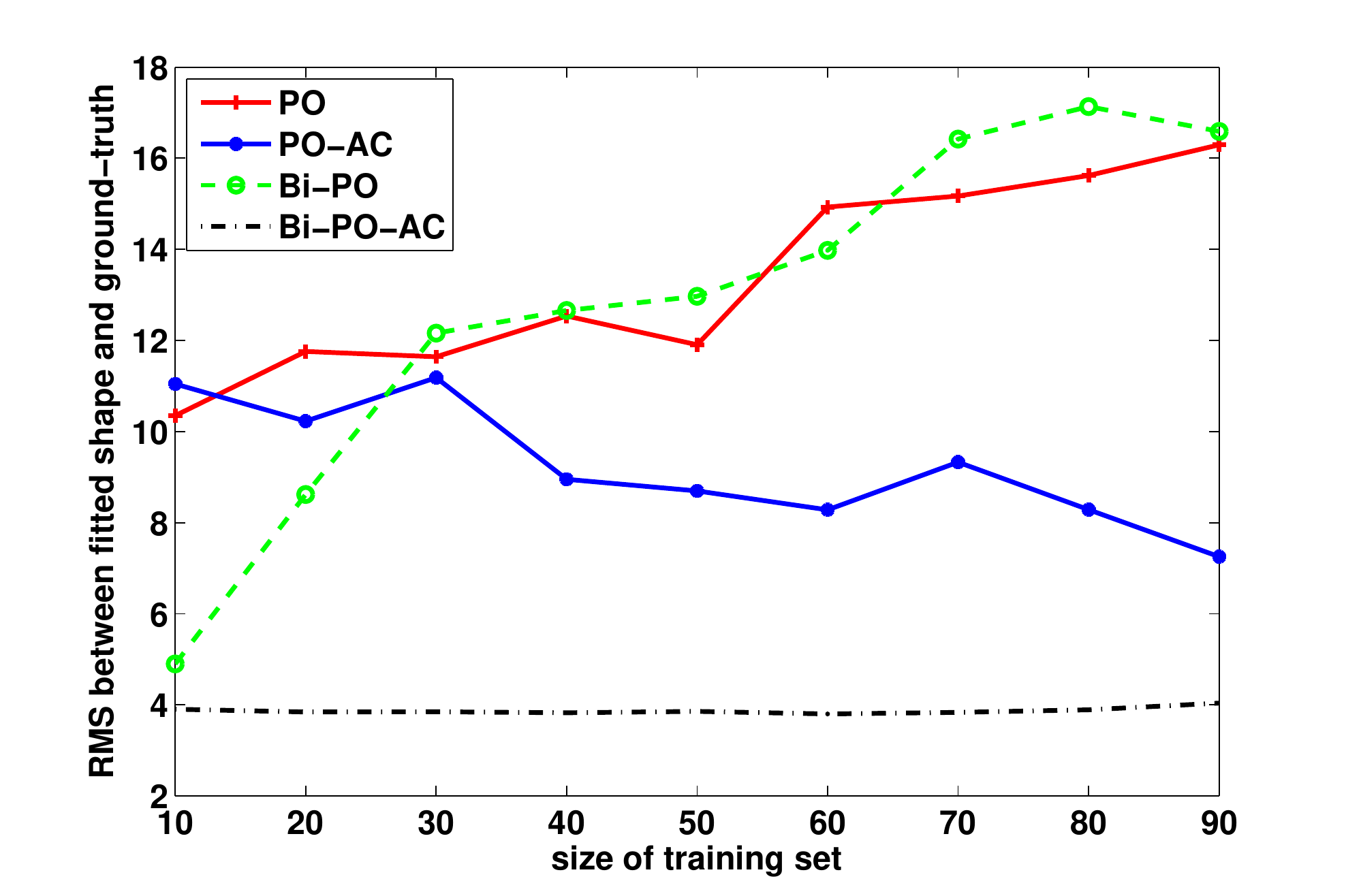}
\vspace{-0.2cm}
\caption{RMSE of fitting for variation of PO.}
\label{fig:RMSPO}
\vspace{-0.3cm}
\end{figure}

\begin{figure}[!t]
\vspace{-0.2cm}
\centering
\includegraphics[width=2.4in]{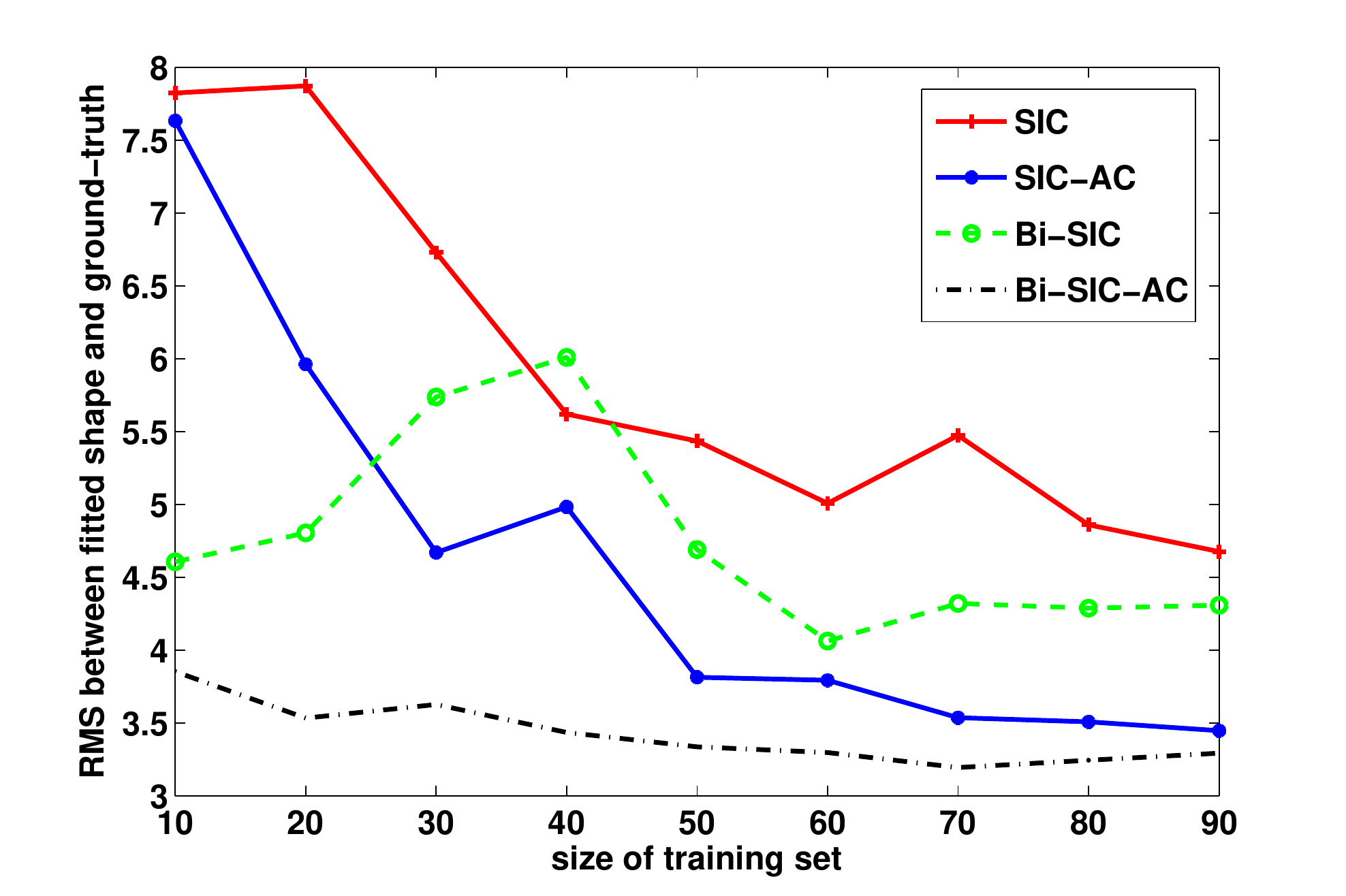}
\vspace{-0.2cm}
\caption{RMSE of fitting for variation of SIC.}
\label{fig:RMSSIC}
\vspace{-0.1cm}
\end{figure}

In our first experiment, we examined the effect of using affine transformation and constraint on both the PO and SIC method as well as the introduced bidirectional warping. Figure \ref{fig:RMSPO} shows the fitting RMSE value of the PO, Bi-PO, PO-AC, and Bi-PO-AC on the frontal subset. Figure \ref{fig:RMSSIC} shows the fitting RMSE value of the SIC, Bi-SIC, SIC-AC, and Bi-SIC-AC on the frontal subset. In both experiments, using affine transformation and having constraint improved the fitting performance. When we have the constraint, it keeps the shape similar to the trained shapes (i.e. face) during the fitting process and prevents the algorithm from resulting non-face shapes. In addition, the affine transformation gives the algorithm more degrees of freedom, and therefore it fits better on unseen samples. It is also shown that bidirectional warping has a better fitting performances than unidirectional warping. Bi-PO and Bi-SIC both have comparative fitting performance and both fit better in comparison with the original unidirectional algorithms.

There are no standard or established choices for the convergence criterion. In this paper, we visually inspected a number of results in the RMSE range of 0-20 and confirmed that those having RMSE less than 5 pixels seem successfully fitted. Figure \ref{fig:subfig_Fitted} shows a sample fitted image having RMSE 4.02.

\begin{figure}[!t]
\vspace{-0.2cm}
\centering
\includegraphics[width=2.4in]{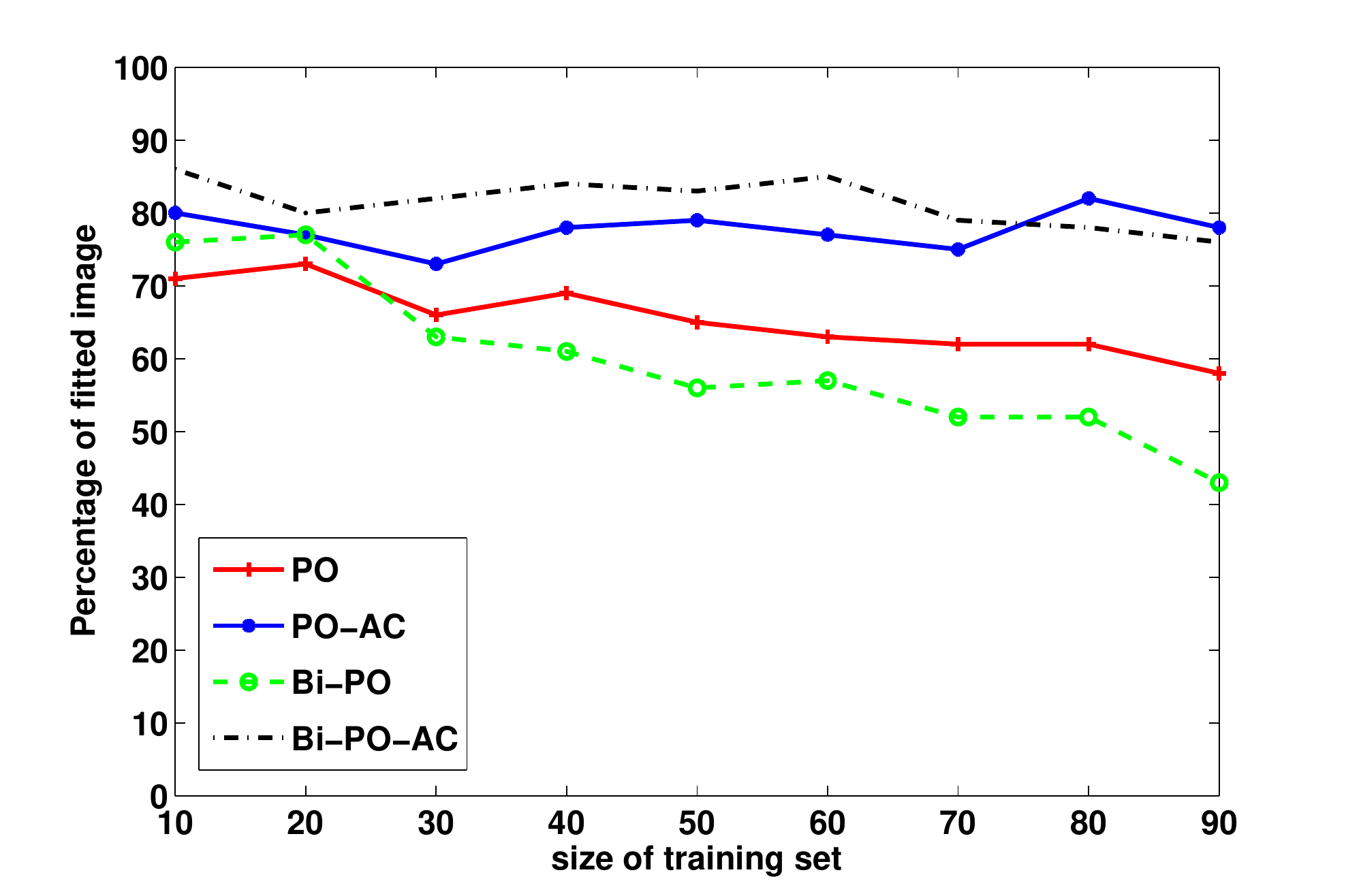}
\vspace{-0.2cm}
\caption{Percentage of fitted images for variation of PO.}
\label{fig:PercentageFittedPO}
\vspace{-0.3cm}
\end{figure}

\begin{figure}[!t]
\vspace{-0.2cm}
\centering
\includegraphics[width=2.4in]{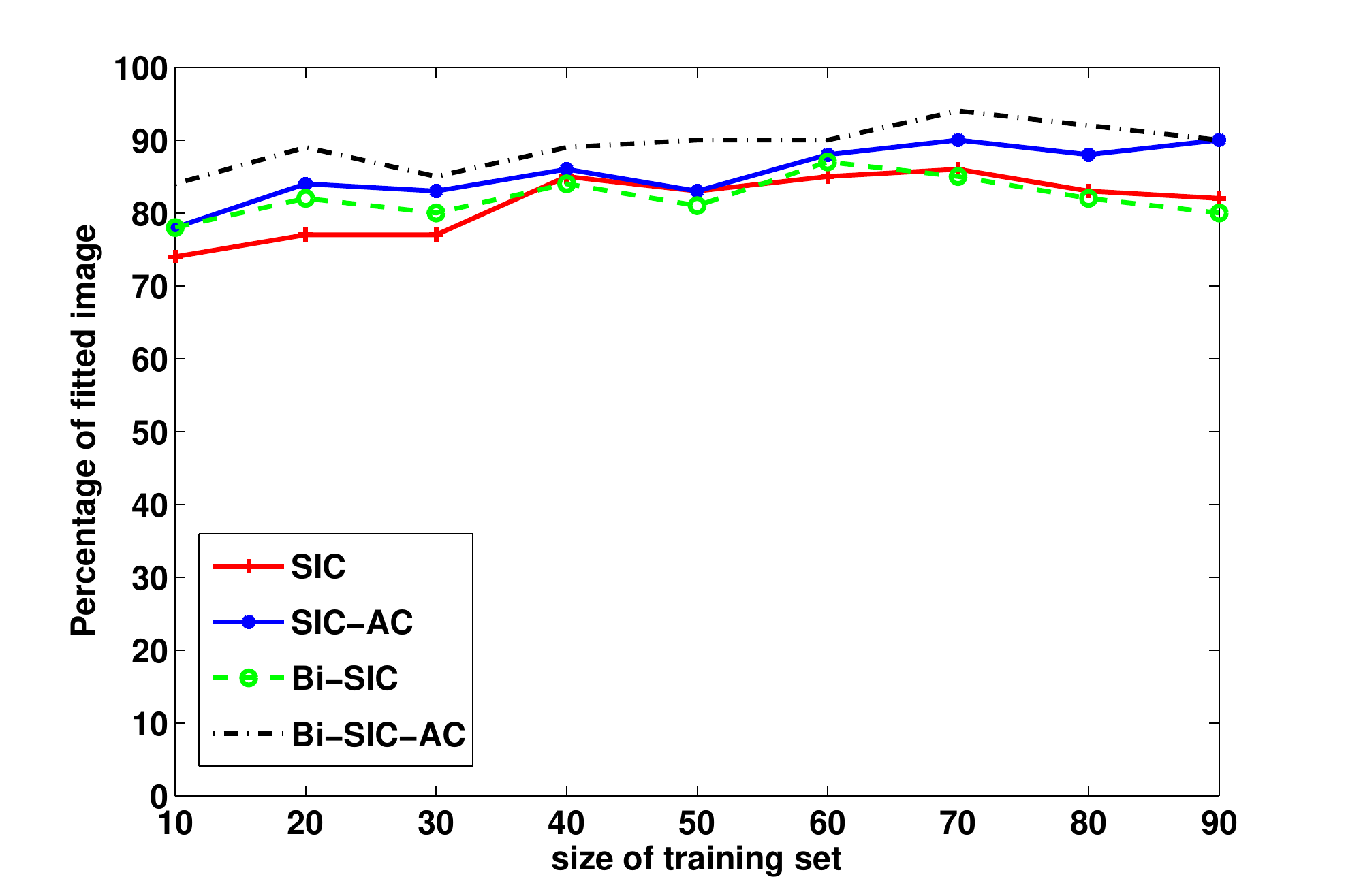}
\vspace{-0.2cm}
\caption{Percentage of fitted images for variation of SIC.}
\label{fig:PercentageFittedSIC}
\vspace{-0.1cm}
\end{figure}

Figure \ref{fig:PercentageFittedPO} shows the percentage of fitted shapes for the frontal subset using PO, Bi-PO, PO-AC, and Bi-PO-AC. Figure \ref{fig:PercentageFittedSIC} shows the percentage of fitted shapes for the frontal subset using SIC, Bi-SIC, SIC-AC and Bi-SIC-AC. As it shown, the bidirectional warping has better performance than the unidirectional method. Also applying the constraint and affine transformation result in a better modeling of unseen images and more convergence on both the PO and SIC. It should be mentioned that the percentage of fitting depends on the threshold value, but empirically both algorithms have more or less similar performance in comparison to each other in a reasonable range of threshold value.

In another experiment, we tested the generalization performance of our proposed approach for different poses. We trained an AAM with 120 images (40 images of each frontal, left and right subsets). To test the generality of the fitting, we fitted the trained model onto the 10 other subjects from each pose. We repeated this experiment five times and averaged the fitting results of the SIC, Bi-SIC, SIC-AC, and Bi-SIC-AC. Initial shape was again the warped average shape obtained from training subjects. Table \ref{tab:RMSFittingLeftRightUpDown_SIC} shows the average RMSE of fitting for frontal, left and right poses. Similarly, we defined a threshold of RMSE less than 5 pixels as the fitted shape. Table \ref{tab:numberOfFittedLeftRightUpDown_SIC} shows the percentage of fitted shapes for frontal, left and right pose subsets. Similar to the previous experiment, using affine transformation and applying constraints on SIC improve the fitting performance. The introduced bidirectional approach also improves the SIC performance significantly, especially when we have pose variations.
 
 \begin{table}[!t]
 \renewcommand{\arraystretch}{1.3}
 \caption{RMSE of fitting on the left and right poses.}
 \label{tab:RMSFittingLeftRightUpDown_SIC}
 \centering
 \begin{tabular}{|l|c|c|c|c|}
 \cline{2-5} \multicolumn{1}{c|}{}  & SIC & SIC-AC & Bi-SIC & Bi-SIC-AC \\
 \hline left & 6.99 & 8.60 & 8.43 & 8.76 \\
 \hline right & 4.07 & 3.40 & 4.02 & 3.37 \\
  \hline frontal & 3.78 & 3.46 & 4.02 & 3.38 \\
 \hline
 \end{tabular}
 \end{table}

 \begin{table}[!t]
 \renewcommand{\arraystretch}{1.3}
 \caption{Percentage of fitted on the left and right poses.}
 \label{tab:numberOfFittedLeftRightUpDown_SIC}
 \centering
 \begin{tabular}{|l|c|c|c|c|}
 \cline{2-5} \multicolumn{1}{c|}{}  & SIC & SIC-AC & Bi-SIC & Bi-SIC-AC \\
 \hline left & 72 & 76 & 62 & 72 \\
 \hline right & 80 & 88 & 80 & 90 \\
 \hline frontal & 86 & 90 & 84 & 96 \\
 \hline
 \end{tabular}
 \end{table}
 
\textbf{Computational Complexity:} The bidirectional method introduces an extra computation in every iterations of fitting. If we assume $n$ is the number of warp parameters, $N$ is the number of pixels, and $m$ is the number of top appearance eigenvectors, the complexity of the PO and SIC methods per iteration are $O(nN+n^2)$ and $O((n+m)^2N+(n+m)^3)$, respectively \cite{baker2004lucas}. In the bidirectional approach, we have $k$ parameters for the chosen global transformation, and in every iterations we need to compute: the gradient of the image (step 7) with the complexity of $O(N)$; the Jacobian $\frac{\partial \mbox{N}}{\partial \mbox{q}}$ (step 8) with the complexity of $O(kN)$; the steepest descent images (step 9) with the complexity of $O(kN)$; the Hessian matrix $\mbox{H}_2$ and invert it (step 10) with the complexity of $O(k^2N+k^3)$; and $\Delta \mbox{q}$ with complexity of $O(kN+k)$. 

The complexity overload of the bidirectional approach is $O(k^2N+k^3)$. The numbers $n$ and $m$ depend on the size of the training set and the model dimensionalities. In most AAM implementations, the dimensionalities of the shape and appearance models are chosen by retaining a fixed percentage (typically 95\%) of the variance in the eigenvalues \cite{Gross2005}. In our experimental results, depending on the size of the training set, $n$ varies between $[10,30]$ and $m$ varies between $[12,70]$. For the affine transformation, $k$ is 6. Hence, the complexity of Bi-PO is at least two times greater than PO, and  the complexity of Bi-SIC is greater than SIC. However, this is based on the assumption of having the same constant factor for all steps.

We implemented all algorithms using Matlab on a windows platform. We executed them on a PC with Intel core Duo 3.00 GHz CPU having 4 GB of RAM, where both implementations have the same termination condition, i.e. the algorithm terminates if the shape does not change or continues for 50 iterations at maximum. In practice, the implemented PO and SIC methods take 3 and 8 seconds for each frame, while the execution of the Bi-PO and Bi-SIC-AC methods take 20 and 27 seconds, respectively, 

\section{Discussion and Conclusions}
\label{Discussion_Conclusions}
In summary, unlike previous image alignment approaches for AAM fitting that warp either the input image (e.g. Lucas-Kanade method) or the appearance template (e.g. inverse compositional algorithm), we warp both the input image for the global transformation and the template for the shape parameters in the fitting process. Warping both the input image and the appearance template causes the AAM to consider more appearance variations, and therefore it can fit better on images with different poses and appearances. We showed that the introduced bidirectional approach can be applied on the ``projected out'' and the ``simultaneously inverse compositional'' approaches for AAM fitting. We also proposed using affine transformation with six degrees of freedom instead of 2D similarity and applying a simple constraint to prevent the fitting algorithm from resulting in shapes far from face geometry.

We tested the performance of the proposed approach on Mutli-PIE dataset. We compared the accuracy of our proposed fitting approach with the PO and SIC methods. First, we trained the AAM with different number of training images and tested the fitting accuracy on unseen images. In another experiment, we then compared the accuracy of fitting on images with different poses. Our experimental results showed that warping both the image and the template makes the AAM fitting more generic. In addition, applying affine transformation gives the algorithm more degrees of freedom to model new face instances and the proposed constraint in the fitting iterations prevents resulting in non-face shapes. In conclusion, our method is promising for modeling and tracking facial images of unseen subjects (i.e. generic model) and also when the accuracy of AAM fitting has priority to the execution speed.

{\small
\bibliographystyle{ieee}

}

\end{document}